\documentclass[letterpaper]{article} % DO NOT CHANGE THIS
\usepackage{aaai24}  % DO NOT CHANGE THIS
\usepackage{times}  % DO NOT CHANGE THIS
\usepackage{helvet}  % DO NOT CHANGE THIS
\usepackage{courier}  % DO NOT CHANGE THIS
\usepackage[hyphens]{url}  % DO NOT CHANGE THIS
\usepackage{graphicx} % DO NOT CHANGE THIS
\usepackage{natbib}  % DO NOT CHANGE THIS AND DO NOT ADD ANY OPTIONS TO IT
\usepackage{caption} % DO NOT CHANGE THIS AND DO NOT ADD ANY OPTIONS TO IT
\usepackage{algorithm}
\usepackage{algorithmic}
\usepackage{amsmath}
\usepackage{amsmath}
\usepackage{amssymb}
\usepackage{multirow}
\usepackage{multicol}
\usepackage{booktabs}
\usepackage{makecell}
\usepackage{subfigure}
\usepackage{array, caption, threeparttable}
\usepackage{newfloat}
\usepackage{listings}
\DeclareCaptionStyle{ruled}{labelfont=normalfont,labelsep=colon,strut=off} % DO NOT CHANGE THIS
\floatstyle{ruled}
\newfloat{listing}{tb}{lst}{}
\floatname{listing}{Listing}
\title{Learning Deformable Hypothesis Sampling for Accurate PatchMatch \\ Multi-View Stereo}
\author{
    Hongjie Li\textsuperscript{}\equalcontrib,
    Yao Guo\textsuperscript{}\equalcontrib,
    Xianwei Zheng\textsuperscript{}\thanks{Corresponding author.},
    Hanjiang Xiong\textsuperscript{}
}
\affiliations{
    The State Key Lab. LIESMARS, Wuhan University, China\\
    \{lihongjie,guoyao\_gy,zhengxw,xionghanjiang\}@whu.edu.cn
}

\usepackage{bibentry}
\begin{document}

\maketitle

\begin{abstract}
This paper introduces a learnable Deformable Hypothesis Sampler (DeformSampler) to address the challenging issue of noisy depth estimation for accurate PatchMatch Multi-View Stereo (MVS). We observe that the heuristic depth hypothesis sampling modes employed by PatchMatch MVS solvers are insensitive to (\romannumeral 1) the piece-wise smooth distribution of depths across the object surface, and (\romannumeral 2) the implicit multi-modal distribution of depth prediction probabilities along the ray direction on the surface points. Accordingly, we develop DeformSampler to learn distribution-sensitive sample spaces to (\romannumeral 1) propagate depths consistent with the scene's geometry across the object surface, and (\romannumeral 2) fit a Laplace Mixture model that approaches the point-wise probabilities distribution of the actual depths along the ray direction. We integrate DeformSampler into a learnable PatchMatch MVS system to enhance depth estimation in challenging areas, such as piece-wise discontinuous surface boundaries and weakly-textured regions. Experimental results on DTU and Tanks \& Temples datasets demonstrate its superior performance and generalization capabilities compared to state-of-the-art competitors. Code is available at \url{https://github.com/Geo-Tell/DS-PMNet}.
\end{abstract}

\section{Introduction}

Multi-View Stereo (MVS) aims to reconstruct dense 3D scene geometry from image sequences with known cameras, which has been widely used in robot perception, 3D reconstruction, and virtual reality. MVS is typically treated as a dense correspondence search problem \cite{galliani2015massively}, but many traditional methods have difficulty in achieving reliable matching within the low-texture, specular, and reflective regions. Learning-based MVS has recently attracted interest in solving this problem by introducing global semantic information for robust matching \cite{yao2018mvsnet,zhang2023geomvsnet}. Although achievements have been made, they still face the challenge of bridging the gap between accuracy and efficiency.

\begin{figure*}[htp!]
    \centering 
\includegraphics[width=0.98\linewidth]{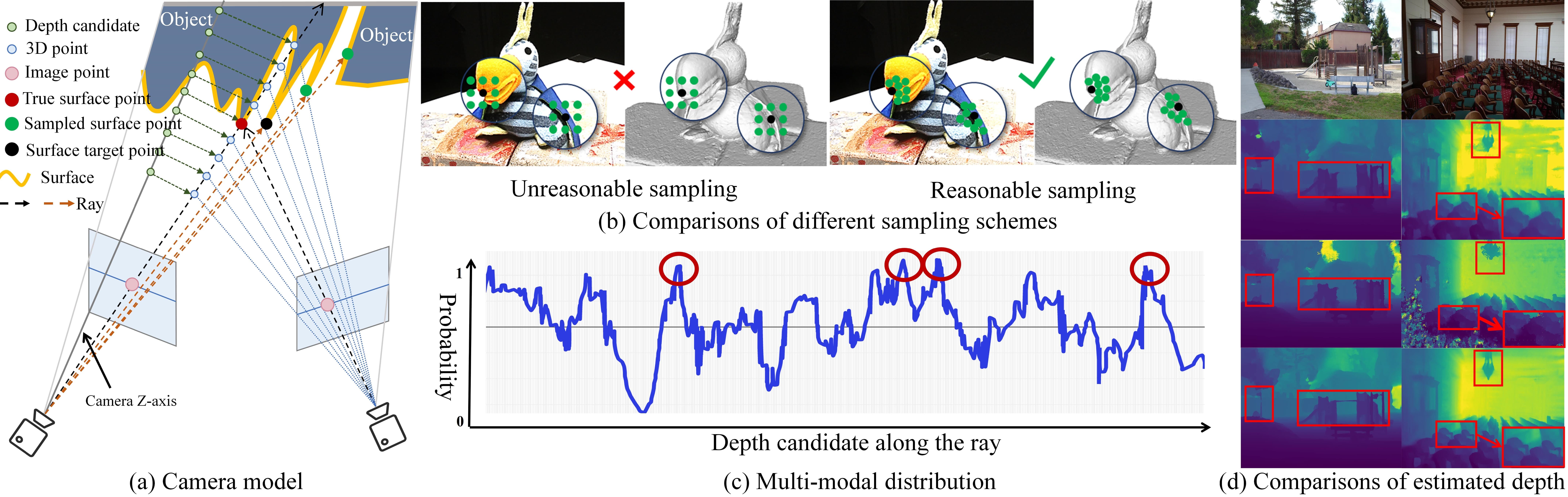} 
    \caption{(a) A camera model is used to understand the hypothesis sampling from the 3D viewpoint. (b) Comparisons between different sampling schemes. The unreasonable sampling template often results in sampling hypotheses from plausible neighboring pixels during propagation. Pixels that appear to be closely adjacent in the image might correspond to significantly distant 3D points in space, and these 3D points could even belong to different objects, like the black and green points in the left sub-figure. (c) An example of the multi-model distribution of depth prediction probabilities illustrates a multi-peak case regarding the minimum cost (or maximum matching probability). (d) Qualitative comparison of depth estimation on the courtroom and playground scenes of the Tanks \& Temples dataset, respectively. From top to bottom: PathchMatchNet \cite{wang2021patchmatchnet}, PatchMatch-RL \cite{lee2021patchmatch} and our DS-PMNet.} 
    \label{figure1} 
\end{figure*}

Learning-based methods commonly involve building a 3D cost volume, followed by a regularization using the 3D CNN for depth regression \cite{yao2018mvsnet}. Consequently, the 3D forms of both cost volume and CNN are undoubtedly restricted by limited resources. To overcome these limitations, many efforts have been made to reduce the cost volume size \cite{gu2020cascade,cheng2020deep} and modify the regularization techniques \cite{yao2019recurrent,yan2020dense}. Recently, a promising solution has emerged, which forgoes the common learning paradigm and re-evolves the traditional PatchMatch MVS into an end-to-end framework, like PatchMatchNet \cite{wang2021patchmatchnet} and PatchMatch-RL \cite{lee2021patchmatch}. These methods follow the idea of patch-based searching and achieve improved results in efficiency and quality. However, we observe that they only transform the traditional pipeline into a trainable one, without adequately considering the implicit depth distribution within scenes for guiding depth hypothesis sampling during depth propagation and perturbation. This flaw directly degenerates the depth estimation performance, as shown in Figure\ref{figure1}(d). Although PatchMatchNet introduces variability to sampling with CNNs, it remains insensitive to the underlying depth distribution. This will hamper the sampling of optimal hypotheses, thereby imposing additional burden on the subsequent learning modules. Therefore, our study raises two crucial questions for hypothesis sampling: (i) \textit{What implicit depth distributions should be learned?} (ii) \textit{How the learned distribution be leveraged to guide hypothesis sampling?}

At the propagation stage, hypotheses of neighboring pixels are sampled to generate a collection for enhancing each pixel's hypothesis space. An implicit piece-wise smooth depth distribution is contained in the depth map due to the scene regularity in the real world. In other words, the depth distribution tends to be smooth within coherent surfaces but can have abrupt shifts between distinct objects or scene elements. However, a preset sampling template is vulnerable when dealing with this implicit distribution \cite{lee2021patchmatch, duggal2019deeppruner}. This unreasonable hypothesis sampling results in a lot of noises with significant hypothesis differences in the collection, thereby causing unstable hypothesis evaluation, as revealed by Figure \ref{figure1}(b). Thus, a well-designed sampling scheme is required to select hypotheses from pixels that align more closely with the object's surface.

At the perturbation stage, fine-grained hypotheses over the scene depth range are expected to be sampled for refining the previously estimated depths. The optimization at this stage has received little attention in recent works. Gipuma \cite{galliani2015massively} employs a bisection strategy to refine sampling, while COLMAP \cite{schonberger2016pixelwise} combines randomly perturbed samples with previous results in various ways. These methods lack the consideration of the uncertainty inherent in previous estimates, leading to redundant and coarse sampling. In this work, we intend to utilize this uncertainty to adaptively adjust the range of perturbations, rather than uniformly sampling for each pixel. In other words, for pixels with high confidence, the sampling should focus on hypotheses closely distributed around the previous estimates. Conversely, the sampling should encompass more dispersed hypotheses for pixels with significant uncertainty to provide a higher likelihood for correcting the estimates. In fact, we find that the cost distribution induced by previous sampling hypotheses offers a good representation of the uncertainty. However, due to the influence of varying imaging conditions, such as lighting, viewpoint, and other factors, this distribution often possesses multi-modal characteristics, as illustrated in Figure \ref{figure1}(c). This means that there is not a single distinct peak representing the lowest cost, which leads to even the true hypothesis failing to derive the lowest cost. 

Based on the discussion above, we develop a learnable Deformable Hypothesis Sampler (DeformSampler) to learn the implicit depth distributions to guide reliable sampling in the learning-based PatchMatch framework. DeformSampler supports each pixel to sample optimal hypotheses at the stages of propagation and perturbation. Two modules are designed to drive this sampler: a plane indicator and a probabilistic matcher. The plane indicator encodes the intra-view feature consistency to learn the implicit depth distribution across the object surface. Using a Laplace Mixture model, the probabilistic matcher models multi-model distribution of depth prediction probabilities along the ray direction. By integrating this sampler into a learning-based PatchMatch framework, we can achieve excellent depth estimation performance, especially in the challenging piece-wise discontinuous surface boundaries and weakly-textured regions. Our method also shows strong generalization ability in both outdoor and indoor scenes, as shown in Figure \ref{figure1}(d).

In summary, our contributions are as follows:
\begin{itemize}
    \item We develop a learnable PatchMatch-based MVS network (\textbf{DS-PMNet}) embedded with DeformSampler to learn implicit depth distribution for guiding the deformable hypothesis sampling.
    \item A plane indicator is designed to capture piece-wise smooth depth distribution for structure-aware depth propagation. 
    \item A probability matcher is designed to model the multi-modal distribution of depth prediction probabilities for uncertainty-aware perturbation.
\end{itemize}

\section{Related Works
}
\subsection{Traditional  MVS}
% Multi-view stereo (MVS) has been studied for decades. 
Traditional MVS methods mainly rely on 3D representations, such as voxel, level-set, polygon mesh, and depth map~\cite{seitz2006comparison} for dense scene geometry reconstruction. Among them, the depth map-based methods usually gain more robust performance for large-scale dense 3D scene recovery by treating MVS as a dense correspondence search problem. In this line of research, PatchMatch MVS is a milestone, which replaces the costly dense point-based search with efficient patch-based search via a random and iterative strategy. Later, the propagation scheme of PatchMatch MVS was optimized for higher efficiency in some works, like Gipuma ~\cite{galliani2015massively} and COLAMP ~\cite{schonberger2016pixelwise}~.

Recently, some methods have promoted the performance of PatchMatch MVS in terms of propagation and evaluation for accurate depth estimation. For propagation, ACMH ~\cite{xu2019multi} adopted an Adaptive Checkerboard Sampling scheme to prioritize good hypotheses. HPM-MVS ~\cite{ren2023hierarchical} enlarged the local propagation region by introducing a Non-local Extensible Sampling Pattern. While these methods provide sensitivity to scene region information, they still have difficulty explicitly capturing such details. To improve hypothesis evaluation in weakly-textured regions, approaches such as utilizing multiscale evaluation strategies, incorporating planar priors \cite{xu2022multi,xu2020planar,romanoni2019tapa}, and employing deformable evaluation regions \cite{wang2023adaptive} have been adopted. In this work, we develop the DeformSampler to effectively impose awareness to scene structures and learn the implicit depth distribution from the current viewpoint for improved hypothesis propagation. 

\subsection{Learning-based MVS}
While traditional solutions perform well in ideal Lambertian scenes, learning-based methods offer better semantic insight and stronger robustness in challenging scenarios. Most learning-based methods were built on MVSNet's foundation. They use warped multi-view features to create cost volumes and adopt 3D CNNs for regularization. Finally, the depths are predicted via regression. Recent works aim to enhance the quality of 3D cost volumes, reduce their size, and refine regularization techniques. To improve quality, techniques like attention mechanisms \cite{lee2022deep,zhu2021multi,cao2022mvsformer}, epipolar-assembling kernels \cite{ma2021epp}, and pixel-wise visibility computation modules \cite{zhang2023vis,xu2020pvsnet} were utilized. For computational efficiency, a common approach is to utilize a coarse-to-fine strategy \cite{gu2020cascade,yang2020cost}, which involves a multi-stage hypothesis sampling. Some variants, like UCS-Net 
\cite{cheng2020deep} and IS-MVSNet \cite{wang2022mvsnet}, adaptively adjust sampling by incorporating uncertainty from depth estimation in earlier stages. For regularization, several studies adopted hybrid 3D U-Net \cite{luo2019p,sormann2020bp}, RNN-CNN \cite{wei2021aa}. In our work, the probabilistic matcher within the DeformSampler employs the same coarse-to-fine strategy but utilizes a more powerful modeling approach to capture the implicit multi-modal distribution of depth prediction probabilities, guiding fine-grained sampling.

\subsection{Learning-based PatchMatch MVS}
Recent advancements have integrated the idea of PatchMatch MVS into end-to-end training frameworks, such as PatchMatchNet \cite{wang2021patchmatchnet} and PatchMatch-RL \cite{lee2021patchmatch}, which have partially bridged the gap between quality and efficiency for learning-based MVS. PatchMatchNet incorporated adaptive propagation and evaluation strategies to achieve efficient depth estimation. PatchMatch-RL argued that traditional methods perform better than learning-based MVS in wide-baseline scenes due to their joint optimization over pixel-wise depth, normals, and visibility estimates. In their follow-up work \cite{lee2022deep}, they further considered the optimization over many high-resolution views and the usage of geometric consistency constraints. Our learnable DS-PMNet is also built upon the PatchMatch MVS but addresses the unreasonable hypothesis sampling issue. The core module, DeformSampler, provides more reliable guidance for propagation and perturbation, leading to a significant performance improvement.

\begin{figure*}[!h]
	\centering
\includegraphics[height=0.295\textwidth]{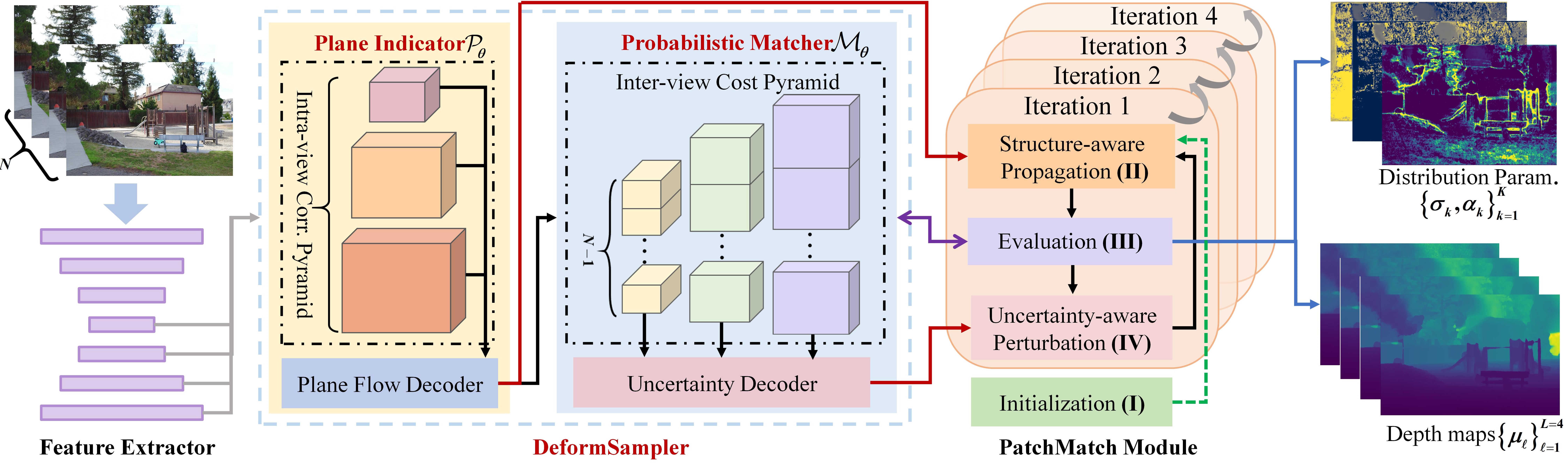}	
	\caption{An overview of the proposed DS-PMNet, which is built upon the PatchMatch framework with the DeformSampler embedded to achieve a deformable hypothesis sampling. DeformSampler learns implicit depth distribution using Plane Indicator $\mathcal P_{\theta}$ and Probability Matcher $\mathcal M_{\theta}$. The purple double sided arrow line indicates that $\mathcal M_{\theta}$ belongs to evaluation.}
\label{figure2}
\end{figure*}

\section{Method}
% This section describes how our DerformSampler guides the PatchMatch solver to sample reliable depth hypotheses based on the implicit depth distributions for accurate depth estimation.
Figure \ref{figure2} shows the entire pipeline of our method. In the following subsections, we first provide an overview of the end-to-end learnable PatchMatch MVS embedded with the DeformSampler (DS-PMNet). Then, we discuss how the two modules (the Plane Indicator $\mathcal P_{\theta}$ and Probability Matcher $\mathcal M_{\theta}$) learn the implicit depth distribution to drive the sampler for implementing deformable depth sampling.

\subsection{Overview}
In the PatchMatch MVS paradigm, each image is sequentially used as a reference image  $I^{r}$, while the remaining images serve as source images $\left \{ I_{i}^{s} \right \}_{i=1}^{N-1}$ to assist in estimating the depth map of $I^{r}$. The estimation process involves stages of initialization, propagation, evaluation, and perturbation, with the latter three stages iterating multiple times. In this work, we perform optimization at four different feature scales, with only one iteration per scale. The detailed DS-PMNet framework is presented in Algorithm 1 of the supplementary material.

We first extract a feature pyramid $ \Psi =\left \{ \varphi_{\ell} \right \}_{\ell=1}^{L}$ for each input image to capture the low-level details and high-level contextual information denoted as follows,
\begin{equation}\label{eq:1}
 \left \{ \varphi ^{r}_{\ell} \right \}_{\ell=1}^{L},\left \{ \varphi ^{s}_{\ell} \right \}_{\ell=1}^{L}=F_{\theta }\left ( I^{r}, I^{s} \right ),
\end{equation}
where $F_{\theta }$ is an encoder, and $\ell \in \left \{ 1,...,L \right \}$ is the indices for the multi-scale features.  In this work, the feature pyramid is constructed with four scales, denoted as $L=4$, corresponding to 1/8, 1/4, 1/2, and 1 of the original image size. To avoid confusion, we only describe four stages of one iteration in the following content, i.e., the subscript $\ell$ is discarded.

In stage \textbf{\uppercase\expandafter{\romannumeral1}}, we randomly initialize a depth map $\text D^{0}$ for $I^{r}$. The known depth range is first divided into $m_{0}$ intervals in the inverse depth space. Then, we randomly sample a depth candidate for each pixel at each interval. This means that each pixel is assigned with $m_{0}$ candidates $\left \{ d_{j} \right \}_{j=1}^{m_{0}}$, which ensures that true hypotheses can be propagated quickly under a limited number of iterations.  

In stage \textbf{\uppercase\expandafter{\romannumeral2}}, the plane indicator $\mathcal P_{\theta}$ guides the structure-aware hypothesis propagation by capturing implicit piece-wise smooth depth distribution of the object's surface. $\mathcal P_{\theta}$ encodes the intra-view feature consistency to estimate a plane flow field $\mathcal{F}$ for $I^{ r}$. For each pixel, $\mathcal{F}$ provides $m_{1}$ neighboring coplanar points to sample hypotheses, resulting in a reliable hypothesis collection $\{ d_j \}_{j=1}^{m_{0}+m_{1}}$.

In stage \textbf{\uppercase\expandafter{\romannumeral3}}, the probabilistic matcher $\mathcal M_{\theta}$ enhances the evaluation of depth candidates in $\{ d_j \}_{j=1}^{m_{0}+m_{1}}$ by modeling implicit multi-modal distribution of depth prediction probabilities, and outputs the prediction uncertainty to guide the subsequent perturbation. $\mathcal M_{\theta}$  first generates a multi-view cost volume $\mathcal{S}=\left \{ \text S_{i} \right \}_{i=1}^{N-1}$, where each element $\text S_{i}$ encodes a matching cost introduced by the depth-induced homography matrix set $\left \{ \text H_{j} \right \}_{j=1}^{m_{0}+m_{1}} $  between $ \varphi ^{r}$ and $ \varphi ^{s}_{i}$. Then, for each pixel in $I^{r}$, the parameter set of Laplace Mixture distribution $\left \{ \psi_{i} \right \}_{i=1}^{N-1} $ is decoded from $\mathcal{S}$ to predict depth map $\text D$ and the corresponding uncertainty map set $\mathcal{U} =\left \{U_{i}\right \}_{i=1}^{N-1}$. 

In stage \textbf{\uppercase\expandafter{\romannumeral4}}, the inferred Laplace Mixture distribution is used to guide the uncertainty-aware perturbation, and a fine-grained hypothesis collection $\{ d_j \}_{j=1}^{m_{2}}$ is sampled. Then, this collection is further input into stage \textbf{\uppercase\expandafter{\romannumeral2}}, and $m_{0}\leftarrow m_{2}$.

\subsection{Plane Indicator for Deformable Propagation}

\begin{figure}[!h]
	\centering
\includegraphics[height=0.26\textwidth]{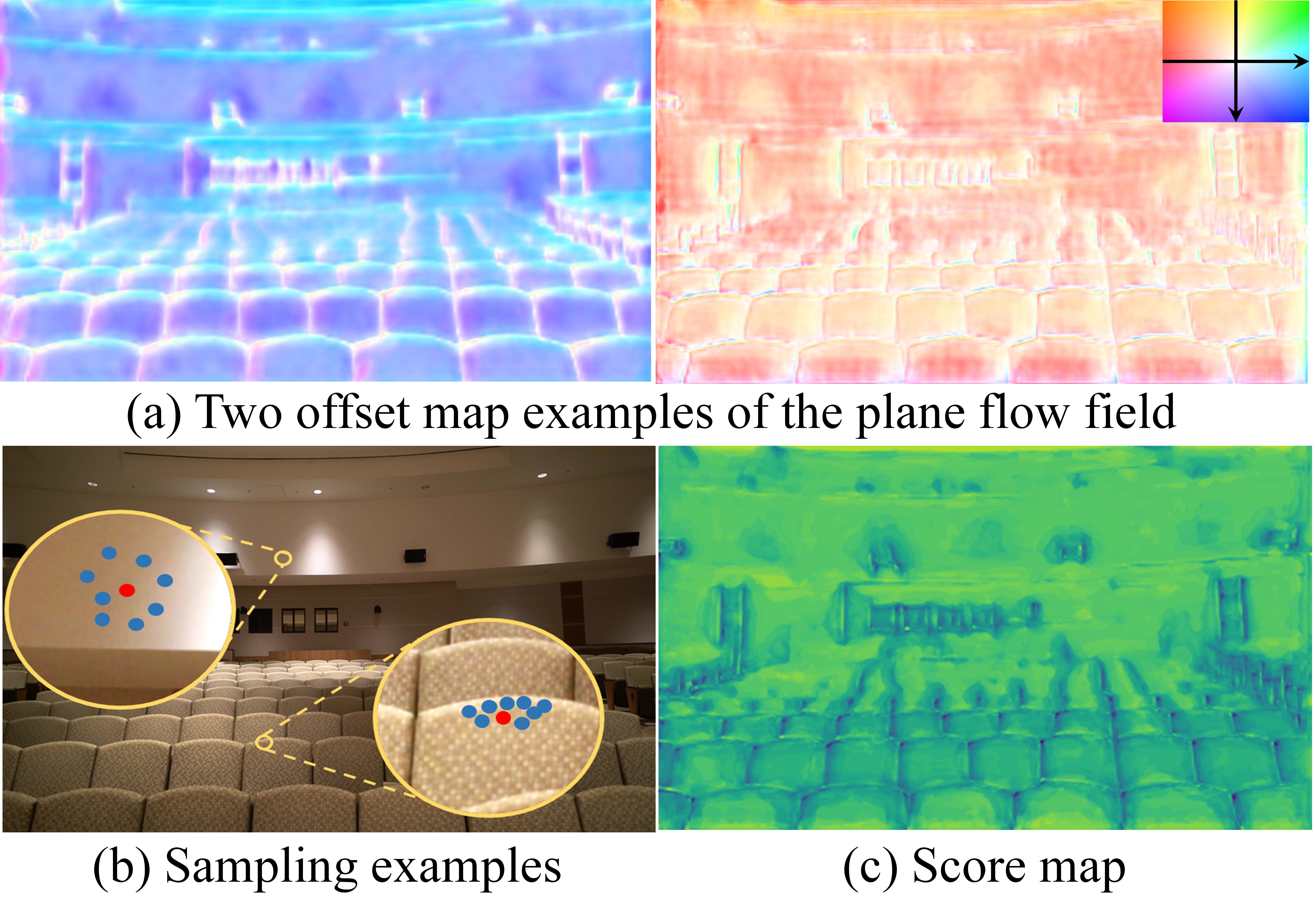}	
	\caption{Visualization of the inferred plane flow field  $\mathcal F$. (a) The offset map is visualized in the form of optical flow, where colors indicate neighboring point directions relative to the current pixel, while color intensity represents offset magnitude. (b) Examples of deformable sampling in the weak texture and object boundary regions. (c) The scene-aligned distribution scoring for the plane flow field.}
	\label{figure3}
\end{figure}

\begin{figure}[!h]
	\centering
\includegraphics[height=0.19\textwidth]{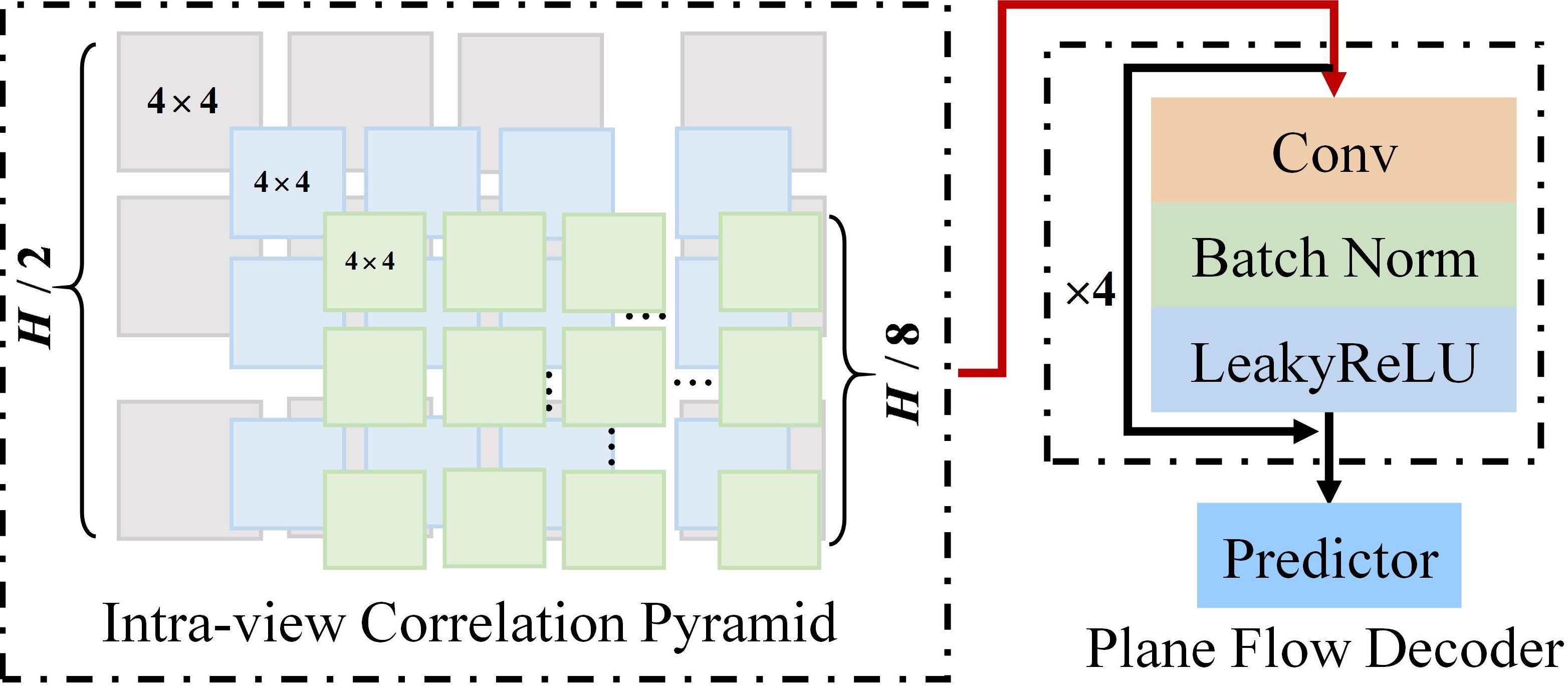}	
	\caption{Illustration of the plane indicator $\mathcal P_{\theta}$. Element in each level of the pyramid is sequentially fed into the plane flow decoder to iteratively refine the plane flow field $\mathcal{F}$. }
	\label{figure4}
\end{figure}

The plane indicator $\mathcal P_{\theta}$ encodes the self-similarity of features within the reference view to learn the relationship between scene structure and depth under the whole PatchMatch solver, thereby decoding a plane flow field $\mathcal{F} \in \mathbb{R}^{H\times W \times 2M}$ that represents the planar regions of the scene. This field contains $M$ offset maps, where each element in the offset map represents the directional displacement between a location and its neighboring points in the same scene plane. Examples of offset maps are shown in Figure \ref{figure3}(a). Utilizing this $\mathcal{F}$, each pixel is guided to sample reliable depth hypotheses from $m_{1}(m_{1}\leq M)$  neighboring points, as shown in Figure \ref{figure3}(b). In general, our $\mathcal P_{\theta}$ consists of two components: an intra-view correlation pyramid $\text C=\left \{ \mathcal C_{\ell} \right \}_{\ell=1}^{L-1}$ and a plane flow decoder $\mathcal D_{\theta}$ in Figure \ref{figure4}.
% inspired by \cite{schonberger2016structure}.
\subsubsection{Intra-view Correlation Pyramid Construction}
Each $\mathcal C_{\ell}$ is generated by calculating the dot product between every pixel in the $\ell_{th}$ feature map and all the pixels within its designated neighborhood. The search radius $R_{1}$ determines the neighborhood range. Specifically, given the feature map $\varphi_{\ell} ^{r}$, each element $c_{\ell}(p,\eta)$ in $\mathcal C_{\ell}\in \mathbb{R}^{H_{\ell}\times W_{\ell}\times R_{1}}$ is defined as 
\begin{equation}\label{eq:2}
c_{\ell}\left ( \text p,\eta \right )=\frac{1}{\sqrt{h_{\ell}}}\left \langle \varphi_{\ell}^{\mathcal R}[\text p],\varphi_{\ell}^{\mathcal R}[\text p+\eta]  \right \rangle,\left \| \eta \right \|_\infty \leq R_{1},
\end{equation}
where  $h_{\ell}$ represents the channel number of $\ell_{th}$ feature map, $\text p \in \mathbb{R}^{2}$ is a coordinate on the feature image and $\eta$ is the offset from this location. The offset is constrained to $\left \| \eta \right \|_\infty \leq R_{1}$. The symbol $\left [ \cdot \right ]$ is used to extract the features at a specific coordinate from the feature map. Each level's search radius $R_{1}$ remains fixed. Therefore, the radius covers the largest feature map area at the top level, gradually reducing with each subsequent level.
% Moreover, before being fed into the decoder, this pyramid is further post-processed by applying channel-wise $L^{2}$-normalization followed by ReLU to effectively reduce the influence of ambiguous elements. 

\subsubsection{Plane Flow Decoder}
This decoder $\mathcal D_{\theta}$ is designed to infer a plane flow field $\mathcal F_{\ell} \in \mathbb{R}^{H_{\ell}\times W_{\ell}\times 2M}$ progressively from the pyramid, which achieves a refined field $\mathcal F$ at last. Inspired by \cite{melekhov2019dgc}, the decoder incorporates four \textit{Conv-BN-LeayReLU} (CBL) blocks and one predictor, as shown in Figure \ref{figure4}.  Dense connections are added among the four blocks to enhance information exchange. Here, a slight adjustment is made in the predictor when inputting elements from different pyramid levels. At level $\ell=1$, the predictor gives a coarse plane flow field $\mathcal F_{1}\in \mathbb{R}^{H_{1}\times W_{1}\times 2M }$. At subsequent levels, the predictor generates a residual $\tilde{\mathcal F}_{\ell} \in \mathbb{R}^{H_{\ell}\times W_{\ell}\times 2 }$ to refine the coarse field further, i.e.,
\begin{equation}\label{eq:3}
\mathcal F_{\ell}\left [ \text p \right ]= \gamma \cdot\mathcal F_{1}^{\uparrow}\left [ \text p \right ]+ \tilde{\mathcal F}_{\ell}\left [ \gamma \cdot\mathcal F_{1}^{\uparrow}\left [ \text p \right ]+ \text p \otimes  \text 1_{M}  \right ],
\end{equation}
where $\text p \in \mathbb{R}^{2}$ is the coordinate on the field, $\gamma$ is the up-sampling factor, $\text 1_{M}$ is a $M\times1$ identity matrix. The symbols $\uparrow$, $\left [ \cdot \right ]$, and $ \otimes$ represent the up-sampling, fetch operation, and Kronecker product operation, respectively.
\begin{figure}[!h]
	\centering
\includegraphics[height=0.26\textwidth]{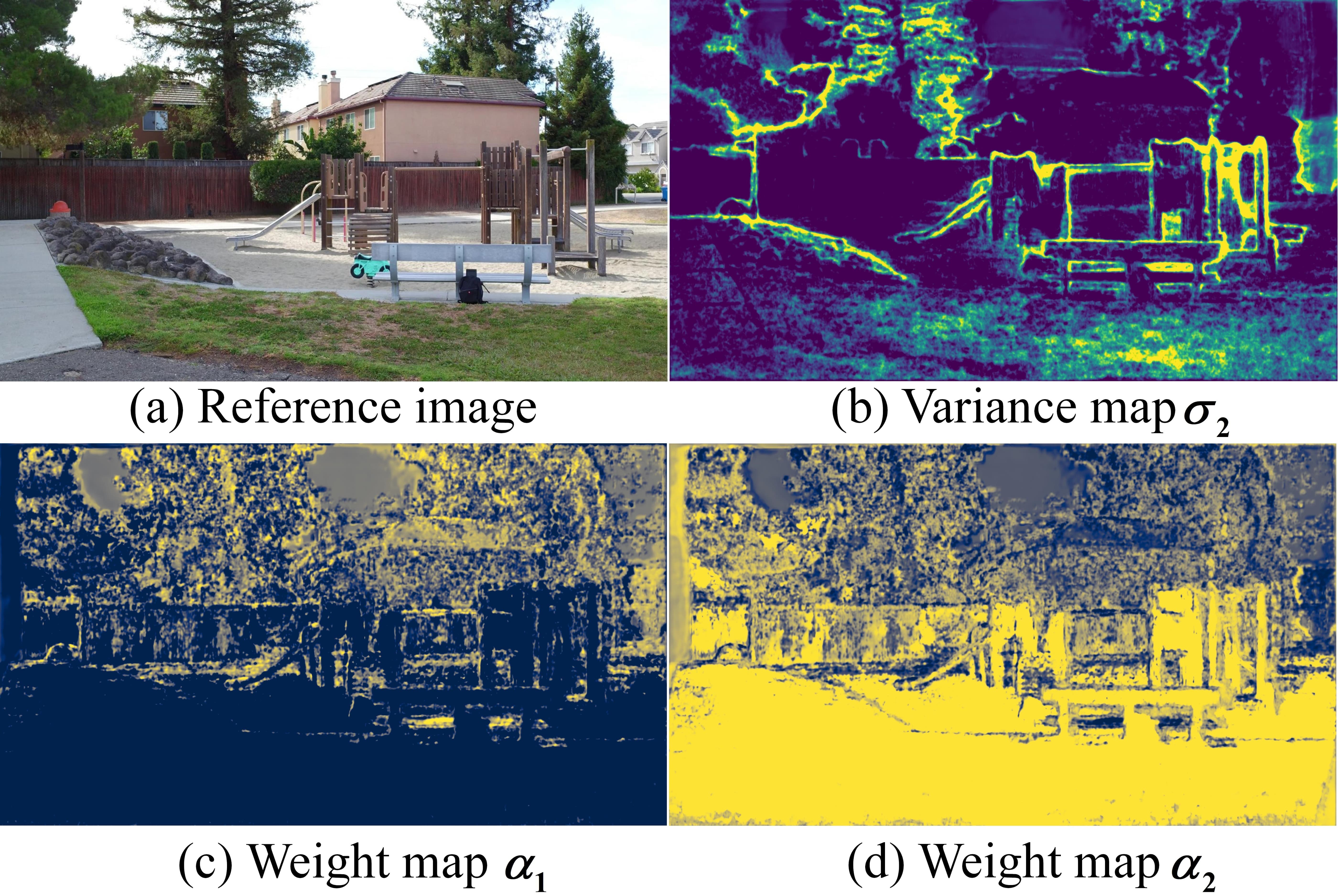}	
	\caption{Visualization of pixel-wise distribution parameters. Brighter colors indicate larger values. Object edge regions exhibit higher uncertainty, i.e., higher $\sigma_{2}$.}
	\label{figure5}
 \end{figure}
\subsection{Probabilistic Matcher for Deformable Perturbation}
The probability matcher $\mathcal M_{\theta}$ is designed to model the multi-modal distribution of the depth prediction probabilities for guiding the fine-grained sampling during perturbation. We adopt a Laplace Mixture model containing two components ($K=2$) to tackle this multi-peak issue, i.e.,
\begin{equation}\label{eq:4}
p\left ( y|\psi \right ) =\alpha _{1}\frac{1}{\sqrt{2}\sigma_{1}}e^{\frac{\sqrt{2}}{\sigma_{1}}\left | y-\mu  \right |}+\alpha _{2}\frac{1}{\sqrt{2}\sigma_{2}}e^{\frac{\sqrt{2}}{\sigma_{2}}\left | y-\mu  \right |},
\end{equation}
where $ \psi=\left \{ \mu,\alpha _{1}, \alpha _{2},\sigma_{1},\sigma_{2} \right \} $ is the distribution parameter set to be estimated, $\alpha _{1}+\alpha _{2}=1$, $y$ is the depth of a specific pixel. The mean $\mu$ of both components is set to be the same to ensure only one peak exists. Additionally, we achieve this by setting $\sigma_{1}$ as a constant for the former to represent the most accurate depth prediction and imposing the constraint $\sigma_{2} > \sigma_{1}>0$ for the latter to model larger errors. Figure \ref{figure5} visualizes an example of pixel-wise distribution parameters. 

 \begin{figure}[h]
	\centering
\includegraphics[height=0.185\textwidth]{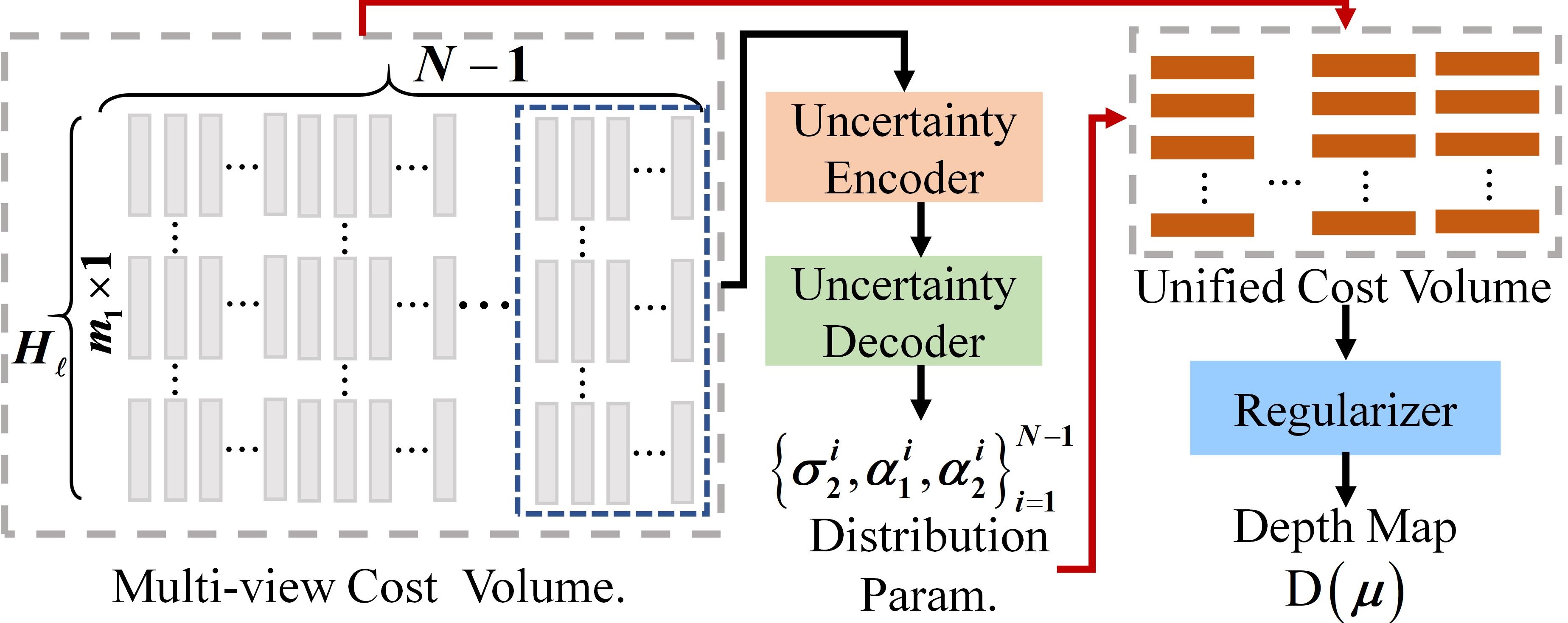}	
	\caption{Illustration of the proposed probabilistic matcher $\mathcal M_{\theta}$. Only a single level of structure is displayed here. }
	\label{figure6}
\end{figure}
\subsubsection{Probabilistic Matcher}
Matcher $\mathcal M_{\theta}$ takes the inter-view cost pyramid as input, where each level of the pyramid contains a multi-view cost volume  $\mathcal S=\left \{ \text S_{i} \right \}_{i=1}^{N-1} $ introduced by the source images.  For each level of the pyramid, the matcher predicts the distribution parameter set $\left \{ \psi_{i} \right \}_{i=1}^{N-1} $ for each pixel in the reference image, representing the matching uncertainty between the reference and different source images. The detailed structure can be seen in Figure \ref{figure6}. This matcher contains two branches. In the first branch, $\mathcal S$ is encoded into an uncertainty embedding, which is then decoded into $\left \{ \sigma^{i}_{2}, \alpha^{i} _{1}, \alpha^{i} _{2}\right \}_{i=1}^{N-1}$ for each pixel. According to the inferred parameters, an uncertainty map set $ \mathcal U=\left \{\text U_{i}\right \}_{i=1}^{N-1}$ between the reference and source images can be obtained by computing depth prediction probabilities on each pixel, i.e., $ u_{i}=P\left ( \left | y-\mu \right |<R_{2}  \right )$, 
\begin{equation} \label{eq:5}
 u_{i}=\alpha^{i} _{1}\left [ 1-e^{ -\sqrt{2}R_{2} }\right ]^{2}+\alpha^{i} _{2}\left [1-e^{ -\sqrt{2}\frac{R_{2}}{\sigma^{i}_{2}} } \right ]^{2},
\end{equation}
where $u_{i}$ is an element of a specific coordinate on $\text U_{i}$, $R_{2}$ is the hyper-parameter determining the acceptable deviation between ground truth and predicted depth map $\mu$. Utilizing those visibility maps $\left \{\text U_{i}  \right \}_{i=1}^{N-1}$, an unified cost volume can be obtained by integrating $\sum_{i=1}^{N-1}\text U_{i}\text S_{i}$. In the second branch, a 3D CNN-based regularizer is adopted to process the unified cost volume to estimate a weighted depth map $D(\mu)$ with the sampled depth hypotheses. More details can be seen in the supplementary.

\subsubsection{Probabilistic Perturbation}
$\sigma _{2}$ is adopted to guide hypothesis sampling because it represents high matching uncertainty. We first integrate the variances from different views into a unified value $E\left ( \sigma _{2} \right )=\sum ^{N-1}_{i=1} u_{i}\sigma _{2}^{i}$. Then, a sampling space is defined as $\left [ \mu \pm \varepsilon E\left ( \sigma _{2} \right ) \right ]$ for each pixel, where $\varepsilon$ is the hyper-parameter. Then, we divide this range into $m_{2}$ bins, each containing an equal portion of probability mass. This ensures that pixels with low uncertainty sample candidates are closer to $\mu$ while pixels with high uncertainty sample more dispersed candidates to rectify $\mu$. Subsequently, we sample the midpoint of each bin as a potential depth candidate. Thus, $j_{th}$ depth candidate is defined as 
\begin{equation} \label{eq:6}
\begin{split}
d_{j} = \frac{1}{2} &\left[ \Phi\left( \frac{j-1}{m_{2}}\tilde{P} + \frac{P^{*}}{2} \right) +  \Phi\left( \frac{j}{m_{2}}\tilde{P}+ \frac{P^{*}}{2} \right) \right],
\end{split}
\end{equation}
where $\Phi\left ( \cdot  \right )$ is used to transform the cumulative probability into coordinates of the Laplace distribution, $m_{2}$ is the number of bins, $\tilde{P} $ is the probability mass covered by range $\left [ \mu \pm \varepsilon E\left ( \sigma _{2} \right ) \right ]$, $P^{*}=1-\tilde{P}$.

\subsection{Loss Function}
We first compute the loss $\mathcal L_{depth} = \sum_{\ell=1}^{L}\left \| \text D^{gt}-\text D_{\ell} \right \|$ between all predicted depth maps $\left \{ \text D_{\ell} \right \}_{\ell=1}^{L}$ and ground truth $\text D^{gt}$. Then, a negative log-likelihood loss $\mathcal L_{NLL}$ is adopted to supervise the fitted mixed Laplace distribution, i.e.,
\begin{equation}\label{eq:7}
\mathcal L_{NLL}=-\frac{1}{N-1} \sum_{\ell=1}^{L}\sum_{i=1}^{N-1}\text{log}p\left ( \text D^{gt}| \psi_{i}  \right ). 
\end{equation}
Thus, the total loss $\mathcal L_{total}$ is defined as $\mathcal L_{total}=\lambda_{1}\mathcal L_{depth}+\lambda_{2}\mathcal L_{NLL}$, where $\lambda_{1}$, $\lambda_{2}$ are the weight factors.

\section{Experiments
}
\subsection{Implementation Setup
}

\subsubsection{Training and Testing} 
Following the previous works like TransMVSNet \cite{ding2022transmvsnet}, we initially train our DS-PMNet on the DTU dataset \cite{jensen2014large} for DTU evaluation. Then, we fine-tune the model on the BlendedMVS dataset \cite{yao2020blendedmvs}, and test it on the Tanks and Temples benchmark \cite{knapitsch2017tanks}. For training, we use 6 DTU images cropped to 512×640 as input in each batch. Our model is trained for 16 epochs with Adam optimizer, starting with a learning rate of 0.001, reduced by 0.2 at epochs 5, 9, and 13. To stabilize training against initial errors from random depth hypotheses, we set the initial learning rate of the probability matcher $\mathcal{M}_{\theta}$ to $10^{-5}$. As for Fine-tuning on BlendedMVS, our model undergoes 10 epochs with an initial learning rate of 0.0002, using 6 input images at a resolution of 576 × 768. The batch size is set to two on NVIDIA RTX 3090 for DTU and one for BlendedMVS.

When assessing the DTU, we use 6 input images at 1152×1600 resolution (N=6). For the Tanks and Temples dataset, N is set to 8, with images at 1024×1920 resolution. We report the results in terms of the accuracy (Acc.), completeness (Comp.), and overall metrics for DTU dataset and evaluate the performance of precision (Pre.), recall (Rec.), and $F_{1}$-score ($F_{1}$) for Tanks and Temples benchmark.

\begin{table}
  \centering
  \resizebox{0.43\textwidth}{!}{
  \def\arraystretch{0.8}%
  \begin{tabular}{c|l|ccc}
    \toprule
    \multicolumn{2}{c}{Methods}  &Acc. $\downarrow$ &Comp. $\downarrow$ &Overall $\downarrow$ \\
    \midrule
    \multirow{2}[0]{*}{\rotatebox{90}{\text{Tra.}}} &
    Gipuma$_{\textit{\tiny{ICCV2015}}}$ & \textbf{0.283} &0.873 &0.578      \\
    & COLMAP$_{\textit{\tiny{CVPR2016}}}$ 
    &0.400 &0.664 &0.532     \\
    \midrule
    \multirow{18}[0]{*}{\rotatebox{90}{ \text{MVSNet}}} &
    MVSNet$_{\textit{\tiny{ECCV2018}}}$   &0.396 &0.527 &0.462      \\
    &
    R-MVSNet$_{\textit{\tiny{CVPR2019}}}$ &0.383 &0.452 &0.417     \\
    &
    Point-MVSNet$_{\textit{\tiny{ICCV2019}}}$ &0.342 &0.411 &0.376  \\
    &
    CasMVSNet$_{\textit{\tiny{CVPR2020}}}$ &0.325 &0.385 &0.355   \\
    &
    CVP-MVSNet$_{\textit{\tiny{CVPR2020}}}$ &0.296 &0.406 &0.351  \\
    &
    UCS-Net$_{\textit{\tiny{CVPR2020}}}$ &0.338 &0.349 &0.344  \\
    &
    AA-RMVSNet$_{\textit{\tiny{ICCV2021}}}$ &0.376 &0.339 &0.357  \\
    &
    UniMVSNet$_{\textit{\tiny{CVPR2022}}}$ &0.352 &0.278 &0.315      \\
    &
    TransMVSNet$_{\textit{\tiny{CVPR2022}}}$ &0.321 &0.289 &0.305      \\
    &
    MVSTER$_{\textit{\tiny{ECCV2022}}}$ &0.350 &0.276 &0.313  \\
    &
    IS-MVSNet$_{\textit{\tiny{ECCV2022}}}$ &0.355 &0.351 &0.359      \\    
    &
    RA-MVSNet$_{\textit{\tiny{CVPR2023}}}$ &0.326 &0.268 &0.297  \\
    &
    DispMVS$_{\textit{\tiny{AAAI2023}}}$ &0.354 &0.324 &0.339  \\
    &
    EPNet$_{\textit{\tiny{AAAI2023}}}$ &0.299 &0.323 &0.311  \\
    &
    GeoMVSNet$_{\textit{\tiny{CVPR2023}}}$ &0.331 &0.259 &0.295  \\
    &
    DMVSNet$_{\textit{\tiny{ICCV2023}}}$ &0.338 &0.272 &0.313  \\
    \midrule
    \multirow{2}[0]{*}{\rotatebox{90}{ \text{PM}}} &
    PatchMatchNet$_{\textit{\tiny{CVPR2021}}}$ &0.427 &0.277 &0.352 \\
    &
    \textbf{Ours} &0.323  &\textbf{0.257}  & \textbf{0.290} \\
    \bottomrule
  \end{tabular}}
    \caption{Quantitative results of reconstructed point clouds
on DTU testing set by using the distance metric $\left [mm  \right ]$(lower is better). Comparison methods are divided into three categories: Traditional approaches (Tra.), MVSNet variants, and PatchMatch series (PM). Acc. and Comp. stand for accuracy and completeness, respectively.}
    \label{table1}
\end{table}
\begin{figure}[!h]
	\centering
\includegraphics[height=0.48\linewidth]{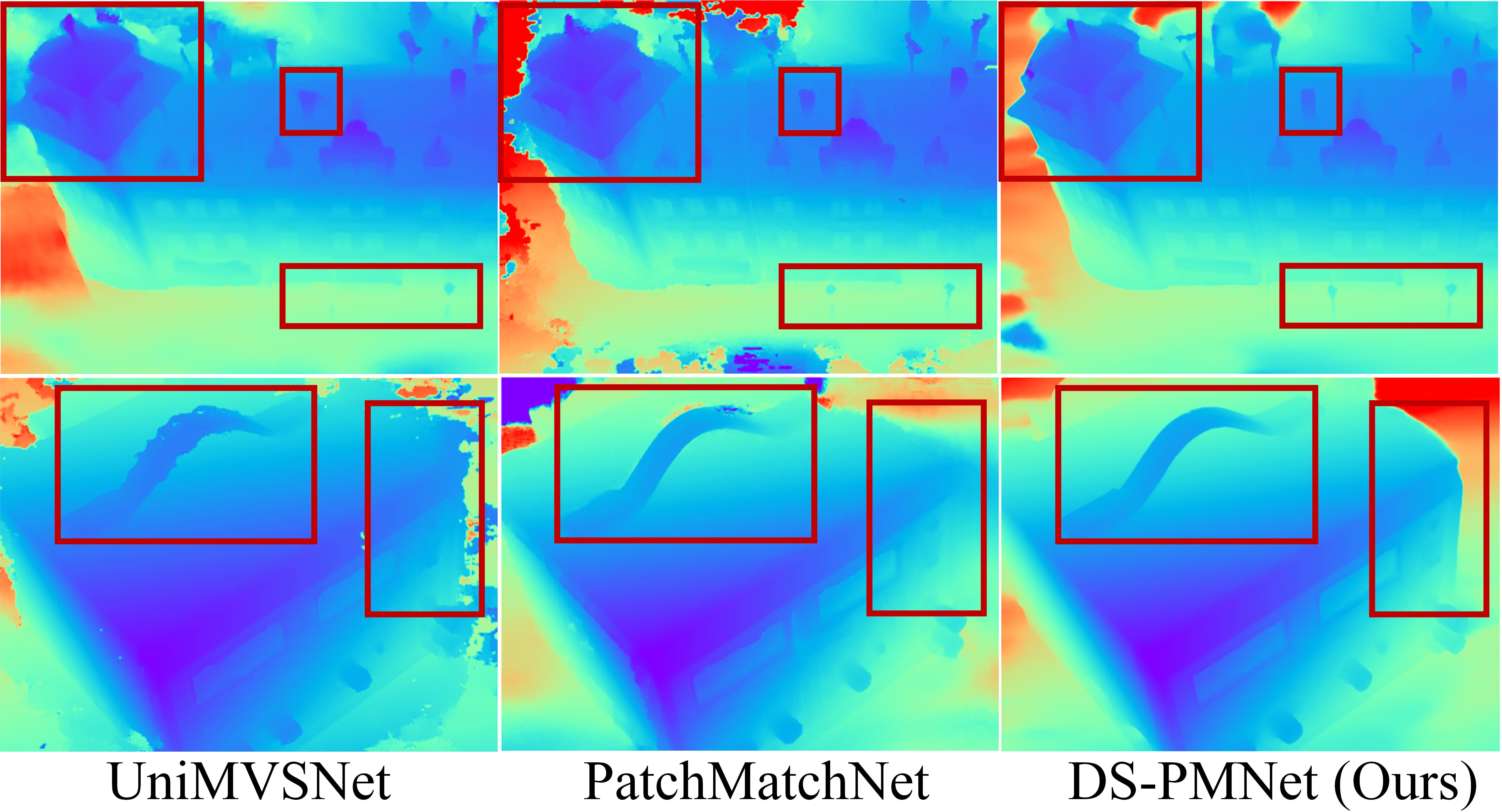}	
	\caption{Qualitative results on DTU testing set. Our model stands out on the edges and thin structures of the depth map, as highlighted by the red box.}
	\label{figure7}
\end{figure}

\subsection{Benchmark Performance
}
\subsubsection{Results on DTU
}
We first evaluate our DS-PMNet on the DTU testing set, where the model is only trained on the DTU dataset. The quantitative results are reported in Table \ref{table1}. Our DS-PMNet outperforms traditional and learning-based methods in the overall metric, achieving a highest score of 0.290. In terms of the accuracy, Gipuma \cite{galliani2015massively} achieves the best results, while our approach demonstrates state-of-the-art performance for completeness. Additionally, we compare our method with PatchMatchNet \cite{wang2021patchmatchnet} and UniMVSNet \cite{peng2022rethinking} for estimated depth maps. As shown in Figure \ref{figure7}, our method excels in recovering the depth of thin structures and object boundaries, where the edges align better with object boundaries than other methods.
\begin{table}
  \centering
  \resizebox{0.45\textwidth}{!}{
  \def\arraystretch{0.8}%
  \setlength\tabcolsep{2.5pt}%
  \begin{tabular}{c|l|ccc|ccc}
    \toprule
    \multicolumn{2}{c}{\multirow{2}[2]{*}{Methods}} &
    \multicolumn{3}{c}{Intermediate ($\uparrow$)}& \multicolumn{3}{c}{Advanced ($\uparrow$)}\\
    \cmidrule(r){3-5}  \cmidrule(r){6-8}
    \multicolumn{2}{c}{} &Pre.  &Rec. &\multicolumn{1}{c}{$F_1$} &Pre. &Rec. &$F_1$\\
    \midrule
    \multirow{2}[0]{*}{\rotatebox{90}{\text{Tra.}}} &
    COLMAP$_{\textit{\tiny{CVPR2016}}}$   &43.16 &44.48 &42.14   &33.65 &23.96 &27.24   \\
    &
    ACMMP$_{\textit{\tiny{TPAMI2023}}}$ & 53.28 &68.50 &59.38   &33.79  & 44.64&37.84   \\

    \midrule
    \multirow{5}[0]{*}{\rotatebox{90}{\small \text{MVSNet}}} &
    MVSNet$_{\textit{\tiny{ECCV2018}}}$   &40.23 &49.70 &43.48 &- &- &-      \\
    &
    CasMVSNet$_{\textit{\tiny{CVPR2020}}}$ &47.62 &74.01 &56.84  &29.68 &35.24 &31.12  \\
    &
    IterMVS$_{\textit{\tiny{CVPR2022}}}$ &46.82 &73.50 &56.22 &28.04 &42.60 &33.24  \\
    &
    Effi-MVS$_{\textit{\tiny{CVPR2022}}}$ &47.53 &71.58 &56.88 &32.23 &41.90 &34.39\\
    &
    EPNet$_{\textit{\tiny{AAAI2023}}}$ &53.26 &71.60 &60.46 &30.75 &44.12 &35.80\\
    \midrule
    \multirow{8}[0]{*}{\rotatebox{90}{\small \text{MVSNet*}}} &
    EPP-MVSNet$_{\textit{\tiny{ICCV2021}}}$ &53.09 &75.58 &61.68    &\textbf{40.09} &34.63 &35.72  \\
    &
    IS-MVSNet$_{\textit{\tiny{ECCV2022}}}$ &55.62 &74.49 &62.82    & 37.03&35.13 &34.87  \\
    % &
    % UniMVSNet_{\textit{\tiny{CVPR2022}}} &57.54 & 73.82&64.36    &33.76 & 47.22&38.96   \\
    &
    RayMVSNet$_{\textit{\tiny{CVPR2022}}}$ &53.21 &69.21 &59.48   &- &- &-   \\
    &
    IterMVS$_{\textit{\tiny{CVPR2022}}}$ &47.53 &74.69 &56.94   &28.70 &44.19 &34.17   \\
    &
    TransMVSNet$_{\textit{\tiny{CVPR2022}}}$ &55.14 &76.73 &63.52 &33.84 &44.29 &37.00     \\
    % &
    % MVSTER_{\textit{\tiny{ECCV2022}}} &51.17 &77.50 &60.92 &33.23 &45.90 &37.53 \\
    % &
    % RA-MVSNet_{\textit{\tiny{CVPR2023}}} &58.68 &75.23 &65.72 &35.00 &47.57 &39.93  \\
    &
    DispMVS$_{\textit{\tiny{AAAI2023}}}$ &49.93 &73.37 & 59.07&26.37 &\textbf{53.67} &34.90 \\
    &
    EPNet$_{\textit{\tiny{AAAI2023}}}$ &\textbf{57.01} &72.57 &63.68&34.26 &\underline{50.54} &\textbf{40.52}  \\
    &
    APD-MVS$_{\textit{\tiny{CVPR2023}}}$ &55.58 &75.06&63.64 &33.77 &49.41 &\underline{39.91} \\
    % &
    % GeoMVSNet_{\textit{\tiny{CVPR2023}}} &59.75 &74.28&65.89 &37.56 &47.74&41.52  \\
    % &
    % DMVSNet_{\textit{\tiny{ICCV2023}}} &53.85 &82.84&64.66 & 34.56& 52.66 &41.17  \\
    \midrule
    \multirow{6}[0]{*}{\rotatebox{90}{\small \text{PM}}} &
    PatchMatchNet$_{\textit{\tiny{CVPR2021}}}$ &43.64 &69.37 &53.15 &27.27 &41.66 &32.31\\
    &
    PatchMatch-RL$^{*}_{\textit{\tiny{ICCV2021}}}$ &45.91 &62.30 &51.81 &30.57 &36.73 &31.78 \\
    &
    PatchMatch-RL$^{*+}_{\textit{\tiny{arXiv2022}}}$ &50.48 &63.27 &55.32 &\underline{38.82} &32.35 &33.80 \\
        &
    % ElasticMVS_{\textit{\tiny{NeurIPS2022}}} &49.44 &71.37 &57.88 &36.28 &41.38 &37.81 \\
    % &
    \textbf{Ours} & 48.02 & 76.26 &58.23 & 32.77 & 41.96 &36.03 \\
    &
    \textbf{Ours*} &\underline{56.02}  &\textbf{76.76}  & \textbf{64.16} &34.29  &48.73  & 39.78\\
    \bottomrule
  \end{tabular}}
    \caption{Quantitative results on Tanks and Temples dataset (unit: \%, higher is better). Methods are also
divided into four categories: Traditional approaches (Tra.),  MVSNet variants trained on DTU, MVSNet variants trained or fine-tuned on BlendedMVS (*), and PatchMatch series (PM). PatchMatch-RL$^{+}$ \cite{lee2022deep} is an extension of PatchMatch-RL. Bold represents the best while underlined represents the second-best.}
  \label{table2}
\end{table}
\begin{figure}[!h]
	\centering
\includegraphics[height=0.43\linewidth]{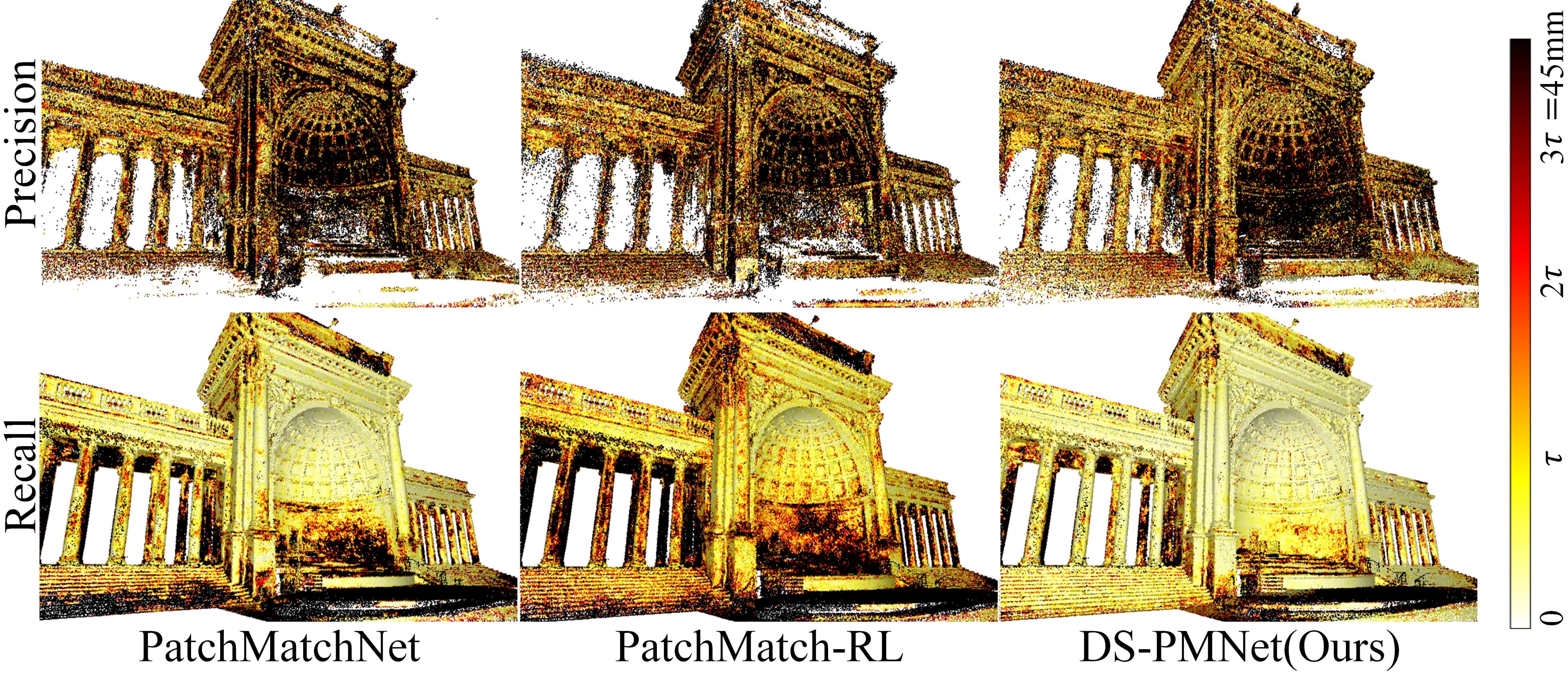}	
	\caption{Qualitative comparison of point cloud reconstruction on the Tanks and Temples benchmark (Dark colors indicate large errors). The distance threshold $\tau$ is scene-dependent and is set to 15mm for the Temple scene.}
	\label{figure8}
\end{figure}

\subsubsection{Results on Tanks and Temples
}
We further validate the generalization ability of our method on the challenging Tanks and Temples dataset. As shown in Table \ref{table2}, our DS-PMNet achieves competitive performance in precision and recall compared to the MVSNet variants. Specifically, our method ranks 1st and 3rd on the intermediate set and advanced set in terms of $F_{1}$ score respectively, outperforming most of methods. Compared with the existing learnable PatchMatch MVS methods, DS-PMNet outperforms better in all metrics. Figure \ref{figure8} provides a qualitative comparison of the reconstructed point clouds for the differen methods, where our method show an enhanced precision and comprehensiveness. The above results demonstrate the robustness and generalization ability of our method.

\subsection{Ablation Studies
}
Ablation Studies are first conducted to independently validate the two modules of DeformSampler (i.e., Plane Indicator $\mathcal{P}_{\theta}$ and Probability Matcher $\mathcal{M}_{\theta}$). Then, comprehensive analysis are made on various $\varepsilon$ settings in the probability matcher to showcase our choice of the parameters.

\noindent \textbf{Effectiveness of DeformSampler}
We first establish a baseline by incorporating Gipuma's fixed sampling modes in the propagation and perturbation stages into our learnable PatchMatch solver. Then, we progressively replace the sampling modes with our proposed plane indicator and probabilistic matcher for validating the effectiveness of our DeformSampler. Quantitative results evaluated on DTU are reported in Table \ref{table3}. The results show that the solver incorporating both modules achieves the highest accuracy and completeness. This demonstrates the capability of our method in effectively learning the underlying depth distributions, guiding reliable hypothesis sampling during the propagation and perturbation stages. Additionally, our baseline outperforms PatchMatchNet in all metrics, demonstrating the superiority of our network design.
\begin{table}
  \label{DTU evaluation}
  \centering
  \resizebox{0.4\textwidth}{!}{
    \begin{tabular}{cccccc}
    \toprule
    & $\mathcal{P}_{\theta}$ & $\mathcal{M}_{\theta}$ & Acc.(mm) & Comp.(mm) & Overall.(mm) \\
    \midrule
    \multirow{1}{*}{}
    & & &0.374 &0.310 &0.342\\
    &\checkmark & &0.368 &0.296 &0.322 \\
    & &\checkmark &0.356 &0.284 &0.320 \\
    &\checkmark &\checkmark &\textbf{0.323} &\textbf{0.257} &\textbf{0.290} \\
    \midrule
    \multicolumn{3}{c}{}{\centering PatchMatchNet} &0.427 &0.277 &0.352 \\
    \bottomrule
    \end{tabular}
  }
  \caption{Ablation studies for the plane indicator $\mathcal{P}_{\theta}$ and probabilistic matcher $\mathcal{M}_{\theta}$ on DTU's testing set (lower is better). }
  \label{table3}
\end{table}

\begin{table}
  \label{DTU evaluation}
  \centering
  \resizebox{0.4\textwidth}{!}{
    \begin{tabular}{ccccc}
    \toprule
    % \multicolumn{1}{c}{} & \multicolumn{3}{c}{}{\centering Mean Distance} \\
     & $\varepsilon$ & Acc.(mm) & Comp.(mm) & Overall.(mm) \\
    \midrule
    &-,1,1,1 &0.345 &0.278 &0.312\\
    &-,2,2,2 &0.352 &0.271 &0.312 \\
    &-,3,2,1 &0.332 &0.267 &0.300 \\
    &-,2,2,1 &\textbf{0.323} &\textbf{0.257} &\textbf{0.290} \\
    \bottomrule
    \end{tabular}
  }
  \caption{Parameter sensitivity testing on DTU for uncertainty-aware perturbation in different stages (lower is better).}
  \label{table4}
\end{table}

\noindent \textbf{Parameter Sensitivity} 
During the perturbation process, the parameter $\varepsilon$ controls the range of the perturbation, i.e., $[\mu-\varepsilon\sigma, \mu+\varepsilon\sigma]$, and different perturbation ranges result in different sampling fineness. Therefore, we constrain the parameters to the subset $\left \{1,2,3  \right \}$ to verify the optimality of our setting. Due to the step-by-step refinement in each iteration, $\varepsilon$ in subsequent iterations must be less than or equal to the previous one. Additionally, the first iteration does not involve the perturbation process. The results of quantitative analysis on DTU are reported in Table \ref{table4}. The best performance is achieved when $\varepsilon$ is set to $\left \{  2, 2, 1\right \}$, followed by $\left \{  3, 2, 1\right \}$. These differences primarily stem from the initial $\varepsilon$ setting. If $\varepsilon$ is set too small ($\varepsilon$=1), the potentially valid hypotheses are excluded, while if too large ($\varepsilon$=3), the redundant noises hamper fine-grained sampling. 
\appendix
\section{Conclusion}
\label{sec:Conclusion}
This paper presents a learnable DeformSampler that is embedded into PatchMatch MVS framework to facilitate the accurate depth estimation in complex scenarios. The proposed DeformSampler can help to sample distribution-sensitive hypothesis space during the propagation and perturbation. Extensive Experiments conducted on several challenging MVS datasets show that DeformSampler can effectively learn the piece-wise smooth depth distribution on the object surface for reliably propagating depth, while successfully capture the multi-modal distribution of depth prediction probabilities to allow for fine-grained hypothesis sampling. Comparisons with existing methods also demonstrate that our method can achieve state-of-the-art performance on MVS benchmarks.

\section{Acknowledgments}
This research was supported by NSFC-projects under Grant 42071370, the Fundamental Research Funds for the Central Universities of China under Grant 2042022dx0001, and Wuhan University-Huawei Geoinformatics Innovation Laboratory.

\bibliography{Accept}

\begin{thebibliography}{35}
\providecommand{\natexlab}[1]{#1}

\bibitem[{Cao, Ren, and Fu(2022)}]{cao2022mvsformer}
Cao, C.; Ren, X.; and Fu, Y. 2022.
\newblock MVSFormer: Multi-View Stereo by Learning Robust Image Features and Temperature-based Depth.
\newblock \emph{Transactions on Machine Learning Research}.

\bibitem[{Cheng et~al.(2020)Cheng, Xu, Zhu, Li, Li, Ramamoorthi, and Su}]{cheng2020deep}
Cheng, S.; Xu, Z.; Zhu, S.; Li, Z.; Li, L.~E.; Ramamoorthi, R.; and Su, H. 2020.
\newblock Deep stereo using adaptive thin volume representation with uncertainty awareness.
\newblock In \emph{Proceedings of the IEEE/CVF Conference on Computer Vision and Pattern Recognition}, 2524--2534.

\bibitem[{Ding et~al.(2022)Ding, Yuan, Zhu, Zhang, Liu, Wang, and Liu}]{ding2022transmvsnet}
Ding, Y.; Yuan, W.; Zhu, Q.; Zhang, H.; Liu, X.; Wang, Y.; and Liu, X. 2022.
\newblock Transmvsnet: Global context-aware multi-view stereo network with transformers.
\newblock In \emph{Proceedings of the IEEE/CVF Conference on Computer Vision and Pattern Recognition}, 8585--8594.

\bibitem[{Duggal et~al.(2019)Duggal, Wang, Ma, Hu, and Urtasun}]{duggal2019deeppruner}
Duggal, S.; Wang, S.; Ma, W.-C.; Hu, R.; and Urtasun, R. 2019.
\newblock Deeppruner: Learning efficient stereo matching via differentiable patchmatch.
\newblock In \emph{Proceedings of the IEEE/CVF international conference on computer vision}, 4384--4393.

\bibitem[{Galliani, Lasinger, and Schindler(2015)}]{galliani2015massively}
Galliani, S.; Lasinger, K.; and Schindler, K. 2015.
\newblock Massively parallel multiview stereopsis by surface normal diffusion.
\newblock In \emph{Proceedings of the IEEE International Conference on Computer Vision}, 873--881.

\bibitem[{Gu et~al.(2020)Gu, Fan, Zhu, Dai, Tan, and Tan}]{gu2020cascade}
Gu, X.; Fan, Z.; Zhu, S.; Dai, Z.; Tan, F.; and Tan, P. 2020.
\newblock Cascade cost volume for high-resolution multi-view stereo and stereo matching.
\newblock In \emph{Proceedings of the IEEE/CVF conference on computer vision and pattern recognition}, 2495--2504.

\bibitem[{Jensen et~al.(2014)Jensen, Dahl, Vogiatzis, Tola, and Aan{\ae}s}]{jensen2014large}
Jensen, R.; Dahl, A.; Vogiatzis, G.; Tola, E.; and Aan{\ae}s, H. 2014.
\newblock Large scale multi-view stereopsis evaluation.
\newblock In \emph{Proceedings of the IEEE conference on computer vision and pattern recognition}, 406--413.

\bibitem[{Knapitsch et~al.(2017)Knapitsch, Park, Zhou, and Koltun}]{knapitsch2017tanks}
Knapitsch, A.; Park, J.; Zhou, Q.-Y.; and Koltun, V. 2017.
\newblock Tanks and temples: Benchmarking large-scale scene reconstruction.
\newblock \emph{ACM Transactions on Graphics (ToG)}, 36(4): 1--13.

\bibitem[{Lee et~al.(2021)Lee, DeGol, Zou, and Hoiem}]{lee2021patchmatch}
Lee, J.~Y.; DeGol, J.; Zou, C.; and Hoiem, D. 2021.
\newblock Patchmatch-rl: Deep mvs with pixelwise depth, normal, and visibility.
\newblock In \emph{Proceedings of the IEEE/CVF International Conference on Computer Vision}, 6158--6167.

\bibitem[{Lee, Zou, and Hoiem(2022)}]{lee2022deep}
Lee, J.~Y.; Zou, C.; and Hoiem, D. 2022.
\newblock Deep PatchMatch MVS with Learned Patch Coplanarity, Geometric Consistency and Adaptive Pixel Sampling.
\newblock \emph{arXiv preprint arXiv:2210.07582}.

\bibitem[{Luo et~al.(2019)Luo, Guan, Ju, Huang, and Luo}]{luo2019p}
Luo, K.; Guan, T.; Ju, L.; Huang, H.; and Luo, Y. 2019.
\newblock P-mvsnet: Learning patch-wise matching confidence aggregation for multi-view stereo.
\newblock In \emph{Proceedings of the IEEE/CVF International Conference on Computer Vision}, 10452--10461.

\bibitem[{Ma et~al.(2021)Ma, Gong, Wang, Huang, Chen, and Yu}]{ma2021epp}
Ma, X.; Gong, Y.; Wang, Q.; Huang, J.; Chen, L.; and Yu, F. 2021.
\newblock Epp-mvsnet: Epipolar-assembling based depth prediction for multi-view stereo.
\newblock In \emph{Proceedings of the IEEE/CVF International Conference on Computer Vision}, 5732--5740.

\bibitem[{Melekhov et~al.(2019)Melekhov, Tiulpin, Sattler, Pollefeys, Rahtu, and Kannala}]{melekhov2019dgc}
Melekhov, I.; Tiulpin, A.; Sattler, T.; Pollefeys, M.; Rahtu, E.; and Kannala, J. 2019.
\newblock Dgc-net: Dense geometric correspondence network.
\newblock In \emph{2019 IEEE Winter Conference on Applications of Computer Vision (WACV)}, 1034--1042. IEEE.

\bibitem[{Peng et~al.(2022)Peng, Wang, Wang, Lai, and Wang}]{peng2022rethinking}
Peng, R.; Wang, R.; Wang, Z.; Lai, Y.; and Wang, R. 2022.
\newblock Rethinking depth estimation for multi-view stereo: A unified representation.
\newblock In \emph{Proceedings of the IEEE/CVF Conference on Computer Vision and Pattern Recognition}, 8645--8654.

\bibitem[{Ren et~al.(2023)Ren, Xu, Zhang, and Yang}]{ren2023hierarchical}
Ren, C.; Xu, Q.; Zhang, S.; and Yang, J. 2023.
\newblock Hierarchical Prior Mining for Non-local Multi-View Stereo.
\newblock \emph{arXiv preprint arXiv:2303.09758}.

\bibitem[{Romanoni and Matteucci(2019)}]{romanoni2019tapa}
Romanoni, A.; and Matteucci, M. 2019.
\newblock Tapa-mvs: Textureless-aware patchmatch multi-view stereo.
\newblock In \emph{Proceedings of the IEEE/CVF International Conference on Computer Vision}, 10413--10422.

\bibitem[{Sch{\"o}nberger et~al.(2016)Sch{\"o}nberger, Zheng, Frahm, and Pollefeys}]{schonberger2016pixelwise}
Sch{\"o}nberger, J.~L.; Zheng, E.; Frahm, J.-M.; and Pollefeys, M. 2016.
\newblock Pixelwise view selection for unstructured multi-view stereo.
\newblock In \emph{Computer Vision--ECCV 2016: 14th European Conference, Amsterdam, The Netherlands, October 11-14, 2016, Proceedings, Part III 14}, 501--518. Springer.

\bibitem[{Seitz et~al.(2006)Seitz, Curless, Diebel, Scharstein, and Szeliski}]{seitz2006comparison}
Seitz, S.~M.; Curless, B.; Diebel, J.; Scharstein, D.; and Szeliski, R. 2006.
\newblock A comparison and evaluation of multi-view stereo reconstruction algorithms.
\newblock In \emph{2006 IEEE computer society conference on computer vision and pattern recognition (CVPR'06)}, volume~1, 519--528. IEEE.

\bibitem[{Sormann et~al.(2020)Sormann, Kn{\"o}belreiter, Kuhn, Rossi, Pock, and Fraundorfer}]{sormann2020bp}
Sormann, C.; Kn{\"o}belreiter, P.; Kuhn, A.; Rossi, M.; Pock, T.; and Fraundorfer, F. 2020.
\newblock Bp-mvsnet: Belief-propagation-layers for multi-view-stereo.
\newblock In \emph{2020 International Conference on 3D Vision (3DV)}, 394--403. IEEE.

\bibitem[{Wang et~al.(2021)Wang, Galliani, Vogel, Speciale, and Pollefeys}]{wang2021patchmatchnet}
Wang, F.; Galliani, S.; Vogel, C.; Speciale, P.; and Pollefeys, M. 2021.
\newblock Patchmatchnet: Learned multi-view patchmatch stereo.
\newblock In \emph{Proceedings of the IEEE/CVF conference on computer vision and pattern recognition}, 14194--14203.

\bibitem[{Wang et~al.(2022)Wang, Gong, Ma, Wang, Zhou, and Chen}]{wang2022mvsnet}
Wang, L.; Gong, Y.; Ma, X.; Wang, Q.; Zhou, K.; and Chen, L. 2022.
\newblock Is-mvsnet: importance sampling-based mvsnet.
\newblock In \emph{European Conference on Computer Vision}, 668--683. Springer.

\bibitem[{Wang et~al.(2023)Wang, Zeng, Guan, Yang, Chen, Liu, Xu, and Luo}]{wang2023adaptive}
Wang, Y.; Zeng, Z.; Guan, T.; Yang, W.; Chen, Z.; Liu, W.; Xu, L.; and Luo, Y. 2023.
\newblock Adaptive Patch Deformation for Textureless-Resilient Multi-View Stereo.
\newblock In \emph{Proceedings of the IEEE/CVF Conference on Computer Vision and Pattern Recognition}, 1621--1630.

\bibitem[{Wei et~al.(2021)Wei, Zhu, Min, Chen, and Wang}]{wei2021aa}
Wei, Z.; Zhu, Q.; Min, C.; Chen, Y.; and Wang, G. 2021.
\newblock Aa-rmvsnet: Adaptive aggregation recurrent multi-view stereo network.
\newblock In \emph{Proceedings of the IEEE/CVF International Conference on Computer Vision}, 6187--6196.

\bibitem[{Xu et~al.(2022)Xu, Kong, Tao, and Pollefeys}]{xu2022multi}
Xu, Q.; Kong, W.; Tao, W.; and Pollefeys, M. 2022.
\newblock Multi-scale geometric consistency guided and planar prior assisted multi-view stereo.
\newblock \emph{IEEE Transactions on Pattern Analysis and Machine Intelligence}, 45(4): 4945--4963.

\bibitem[{Xu and Tao(2019)}]{xu2019multi}
Xu, Q.; and Tao, W. 2019.
\newblock Multi-scale geometric consistency guided multi-view stereo.
\newblock In \emph{Proceedings of the IEEE/CVF Conference on Computer Vision and Pattern Recognition}, 5483--5492.

\bibitem[{Xu and Tao(2020{\natexlab{a}})}]{xu2020planar}
Xu, Q.; and Tao, W. 2020{\natexlab{a}}.
\newblock Planar prior assisted patchmatch multi-view stereo.
\newblock In \emph{Proceedings of the AAAI Conference on Artificial Intelligence}, volume~34, 12516--12523.

\bibitem[{Xu and Tao(2020{\natexlab{b}})}]{xu2020pvsnet}
Xu, Q.; and Tao, W. 2020{\natexlab{b}}.
\newblock Pvsnet: Pixelwise visibility-aware multi-view stereo network.
\newblock \emph{arXiv preprint arXiv:2007.07714}.

\bibitem[{Yan et~al.(2020)Yan, Wei, Yi, Ding, Zhang, Chen, Wang, and Tai}]{yan2020dense}
Yan, J.; Wei, Z.; Yi, H.; Ding, M.; Zhang, R.; Chen, Y.; Wang, G.; and Tai, Y.-W. 2020.
\newblock Dense hybrid recurrent multi-view stereo net with dynamic consistency checking.
\newblock In \emph{European conference on computer vision}, 674--689. Springer.

\bibitem[{Yang et~al.(2020)Yang, Mao, Alvarez, and Liu}]{yang2020cost}
Yang, J.; Mao, W.; Alvarez, J.~M.; and Liu, M. 2020.
\newblock Cost volume pyramid based depth inference for multi-view stereo.
\newblock In \emph{Proceedings of the IEEE/CVF Conference on Computer Vision and Pattern Recognition}, 4877--4886.

\bibitem[{Yao et~al.(2018)Yao, Luo, Li, Fang, and Quan}]{yao2018mvsnet}
Yao, Y.; Luo, Z.; Li, S.; Fang, T.; and Quan, L. 2018.
\newblock Mvsnet: Depth inference for unstructured multi-view stereo.
\newblock In \emph{Proceedings of the European conference on computer vision (ECCV)}, 767--783.

\bibitem[{Yao et~al.(2019)Yao, Luo, Li, Shen, Fang, and Quan}]{yao2019recurrent}
Yao, Y.; Luo, Z.; Li, S.; Shen, T.; Fang, T.; and Quan, L. 2019.
\newblock Recurrent mvsnet for high-resolution multi-view stereo depth inference.
\newblock In \emph{Proceedings of the IEEE/CVF conference on computer vision and pattern recognition}, 5525--5534.

\bibitem[{Yao et~al.(2020)Yao, Luo, Li, Zhang, Ren, Zhou, Fang, and Quan}]{yao2020blendedmvs}
Yao, Y.; Luo, Z.; Li, S.; Zhang, J.; Ren, Y.; Zhou, L.; Fang, T.; and Quan, L. 2020.
\newblock Blendedmvs: A large-scale dataset for generalized multi-view stereo networks.
\newblock In \emph{Proceedings of the IEEE/CVF conference on computer vision and pattern recognition}, 1790--1799.

\bibitem[{Zhang et~al.(2023{\natexlab{a}})Zhang, Li, Luo, Fang, and Yao}]{zhang2023vis}
Zhang, J.; Li, S.; Luo, Z.; Fang, T.; and Yao, Y. 2023{\natexlab{a}}.
\newblock Vis-mvsnet: Visibility-aware multi-view stereo network.
\newblock \emph{International Journal of Computer Vision}, 131(1): 199--214.

\bibitem[{Zhang et~al.(2023{\natexlab{b}})Zhang, Peng, Hu, and Wang}]{zhang2023geomvsnet}
Zhang, Z.; Peng, R.; Hu, Y.; and Wang, R. 2023{\natexlab{b}}.
\newblock GeoMVSNet: Learning Multi-View Stereo With Geometry Perception.
\newblock In \emph{Proceedings of the IEEE/CVF Conference on Computer Vision and Pattern Recognition}, 21508--21518.

\bibitem[{Zhu et~al.(2021)Zhu, Peng, Li, Shen, Zhang, and Lei}]{zhu2021multi}
Zhu, J.; Peng, B.; Li, W.; Shen, H.; Zhang, Z.; and Lei, J. 2021.
\newblock Multi-view stereo with transformer.
\newblock \emph{arXiv preprint arXiv:2112.00336}.

\end{thebibliography}
\end{document}